\documentclass{llncs}

\usepackage{graphicx}
\usepackage{mathrsfs}
\usepackage{amsmath,bm}
\usepackage{algorithm,algorithmic}
\usepackage{amsfonts}
\usepackage[colorlinks=true, citecolor=blue, urlcolor=black, linkcolor=red]{hyperref}
\usepackage{color}

\usepackage{llncsdoc}

\title{Unsupervised Diverse Colorization via Generative Adversarial Networks}
\author{Yun Cao, Zhiming Zhou, Weinan Zhang, Yong Yu}
\institute{Shanghai Jiao Tong University, Shanghai, China\\
\email{\{yuncao, heyohai, wnzhang, yyu\}@apex.sjtu.edu.cn}}

\begin{document}
	
\maketitle
	
\begin{abstract}
Colorization of grayscale images is a hot topic in computer vision. Previous research		
mainly focuses on producing a color image to recover the original one in a supervised learning fashion.
However, since many colors share the same gray value, an input grayscale image could be diversely
colorized while maintaining its reality.
In this paper, we design a novel solution for unsupervised diverse colorization. Specifically,
we leverage conditional generative adversarial networks to model the distribution of real-world item colors,
in which we develop a fully convolutional generator
with multi-layer noise to enhance diversity, with multi-layer condition concatenation to maintain reality, 
and with stride 1 to keep spatial information. With such a novel network architecture,
the model yields highly competitive performance on the open LSUN bedroom dataset.
The Turing test on 80 humans further indicates our generated color schemes
are highly convincible.
\end{abstract}

\section{Introduction}

Image colorization assigns a color to each pixel of a target grayscale image. Early colorization methods \cite{Levin-et-al:ACM-2004,Qu-et-al:ACM-2006} require users to provide considerable scribbles on the grayscale image,
which is apparently time-consuming and requires expertise. Later research provides more automatic colorization methods.
Those colorization algorithms differ in the ways of how they model the correspondence between grayscale and color.

Given an input grayscale image, non-parametric methods
first define one or more color reference images (provided by a human or retrieved
automatically) to be used as source data. Then, following the Image Analogies
framework \cite{Hertzmann-et-al:SIGGRAPH-2001}, the color is transferred onto the input image from analogous regions
of the reference image(s) \cite{Welsh-et-al:SIGGRAPH-2002,Liu-et-al:ACM-2008,Gupta-et-al:ACM-2012,Chia-et-al:ACM-2011}.
Parametric methods, on the other hand, learn prediction functions from large datasets of color images in the training stage, posing
the colorization problem as either regression in the continuous color space \cite{Cheng-et-al:ICCV-2015,Deshpande-et-al:ICCV-2015,Zhang-et-al:ECCV-2016}
or classification of quantized color values \cite{Charpiat-et-al:ECCV-2008}, which is a supervised learning fashion.

\begin{figure}[t]
	\centering
	\includegraphics[width=0.6\textwidth]{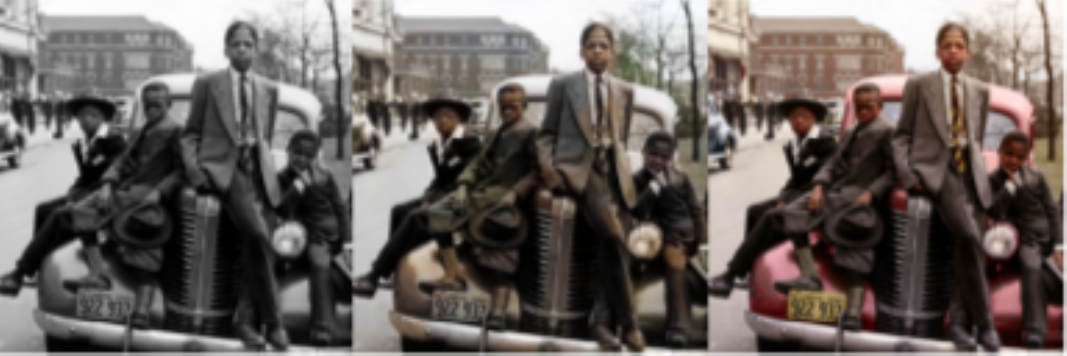}
	\caption{Left: The original grayscale image.
	Middle: Image colorized by non-adversarial CNNs.
	Right: Image colorized by human on Reddit.
	(Figure from \cite{Stephen:2016})}
	\label{fig:brown}
\end{figure}

Whichever seeking the reference images or learning a color prediction model,
all above methods share a common goal, i.e. to provide a color image closer to the original one.
But as we know, many colors share the same gray value. Purely from a grayscale image, one cannot tell what color of clothes a girl is wearing or what color a bedroom wall is. Those methods all produce a deterministic mapping function. Thus when an item could have diverse colors, their models tend to provide a weighted average brownish color as pointed out in \cite{Stephen:2016} (See Figure \ref{fig:brown} as an example).

In this paper, to avoid this sepia-toned colorization, we use conditional generative adversarial networks (GANs) \cite{goodfellow-et-al:NIPS-2014}
to generate diverse colorizations for a single grayscale image while maintaining their reality.
GAN is originally proposed to generate vivid images from some random noise. It is composed of two adversarial parts: a generative model $G$
that captures the data distribution, and a discriminative model $D$ that estimates the probability of whether an image is real or generated by $G$.
The generator part tries to map an input noise to a data
distribution closer to the ground truth data distribution, while
the discriminator part tries to distinguish the generated ``fake"
data, which comes to an adversarial situation. By careful designation
of both generative and discriminative parts, the generator will
eventually produce results, forming a distribution very close to
the ground truth distribution, and by controlling the input noise
we can get various results of good reality. 
Thus conditional GAN is a much more suitable framework to handle
diverse colorization than other CNNs.
Meanwhile, as the discriminator only needs the signal of whether a training instance is real or generated,
which is directly provided without any human annotation during the training phase, the task is in an unsupervised learning fashion.

On the aspect of model designation, unlike many other conditional GANs \cite{Isola-et-al:CoRR-2016} using convolution layers as the encoder and deconvolution layers as the decoder, we build a fully convolutional generator and each convolutional layer is splinted by a concatenate layer to continuously render the conditional grayscale information. Additionally, to maintain the spatial information, we set all convolution stride
to 1 to avoid downsizing data. We also concatenate noise channels to the first half convolutional layers of the generator to attain more diversity in the color image generation process.
As the generator $G$ would capture the color distribution, we can alter the colorization result by changing the input noise.
Thus we no longer need to train an additional independent model for each color scheme like \cite{Cheng-et-al:ICCV-2015}.

As our goal alters from producing the original colors to producing realistic diverse colors,
we conduct questionnaire surveys as a Turing test instead of calculating the root mean squared error (RMSE) comparing the original image to measure our colorization result.
The feedback from 80 subjects indicates that our model successfully produces high-reality color images, yielding more than $62.6\%$ positive feedback while the rate of ground truth images is 70.0\%.
Furthermore, we perform a significance $t$-test to compare the percentages of human judges as real color images for each test instance (i.e. a real or generated color image). The resulting $p$-value is $0.1359>0.05$, which indicates that there is no significant difference between our generated color images and the real ones. We share the repeatable experiment code for further research\footnote{Experiment code is available at \url{https://github.com/ccyyatnet/COLORGAN}.}.

\section{Related Work}

\subsection{Diverse Colorization}

The problem of colorization was proposed in the last century, but the research of diverse colorization was not paid much attention until this decade.
In \cite{Cheng-et-al:ICCV-2015}, they used additionally trained model to handle diverse colorization of a scene image particularly in day and dawn.
\cite{Zhang-et-al:ECCV-2016} posed the colorization problem as a classification task and use class re-balancing at
training time to increase the colorfulness of the result.
And in the work of \cite{Deshpande-et-al:CoRR-2016}, a low dimensional embedding of color fields using a variational auto-encoder (VAE) is learned.
They constructed loss terms for the VAE decoder that avoid blurry outputs and take into account the uneven distribution of pixel colors
and finally developed a conditional model for the multi-modal distribution between gray-level image and the color field embeddings.

Compared with above work, our solution uses conditional generative adversarial networks to achieve unsupervised diverse colorization in a generic way with little domain knowledge of the images.

\subsection{Conditional GAN}

Generative adversarial networks (GANs) \cite{goodfellow-et-al:NIPS-2014} have attained much attention in unsupervised learning research during the recent 3 years.
Conditional GANs have been widely used in various computer vision scenarios.
\cite{Reed-et-al:ICML-2016} used text to generate images by applying adversarial networks.
\cite{Isola-et-al:CoRR-2016} provided a general-purpose image-to-image translation model that handles tasks like label to scene, aerial to map, day to night, edges to photo and also grayscale to color.

Some of the above works may share a similar goal with us, but our conditional GAN structure differs a lot from previous work in several architectural choices mainly for the generator. Unlike other generators which employ an encoder-like front part consisting of multiple convolution layers and a decoder-like end part consisting of multiple deconvolution layers, 
our generator uses only convolution layers all over the architecture, and does not downsize
data shape by applying convolution stride no more than 1 and no pooling operation.
Additionally, we add multi-layer noise to generate more diverse colorization, while using multi-layer conditional information to keep the generated image highly realistic. 

\section{Methods}

\subsection{Problem formulation}

\begin{figure}[t]
	\centering
	\includegraphics[clip, trim={365pt, 0, 0, 150pt}, width=0.7\textwidth]{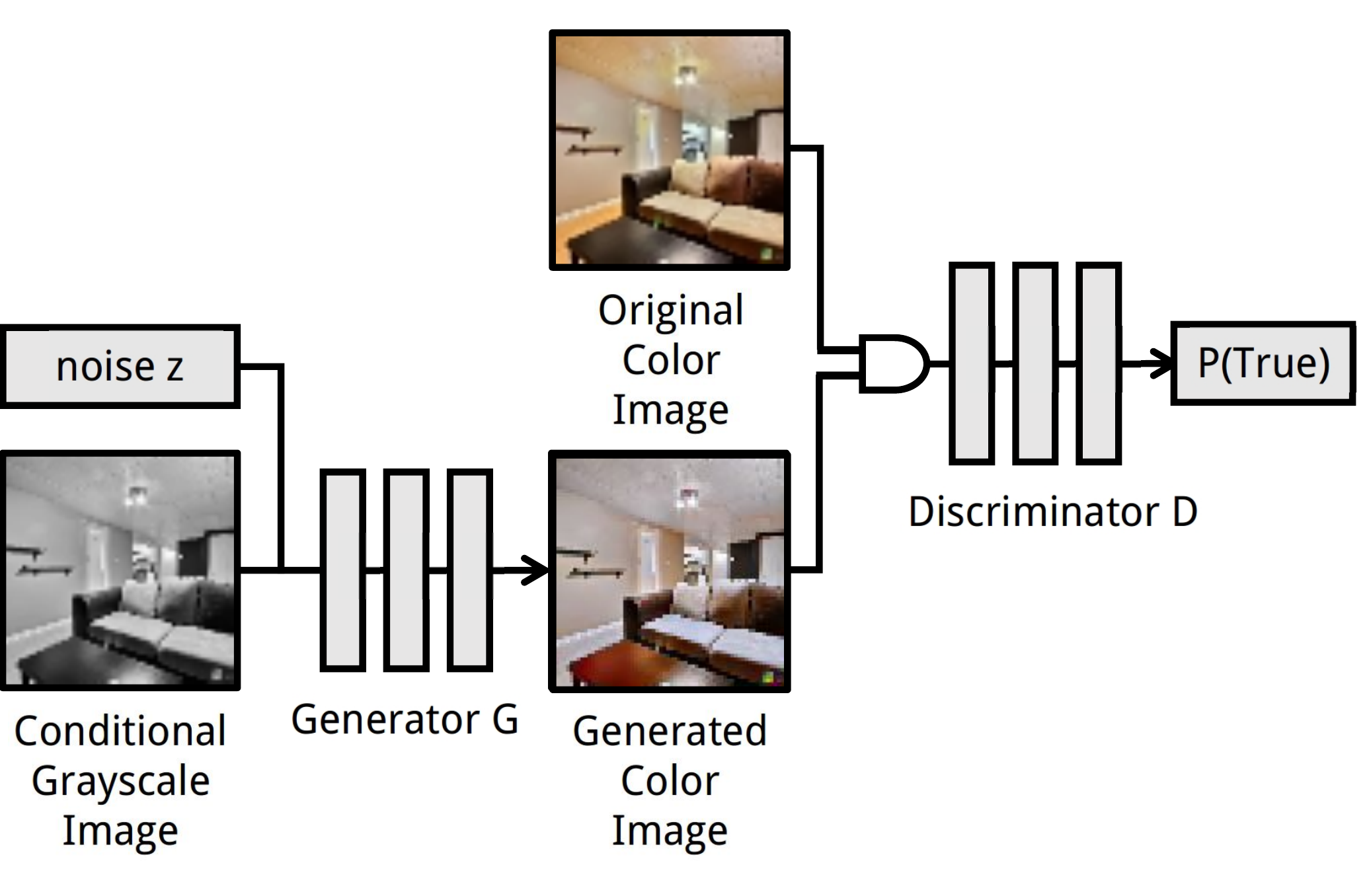}
	\caption{The illustration of conditional GAN.
		Generator $G$ is given conditional information (Grayscale image) together with noise $\bm{z}$,
		and produces generated color channels.
		Discriminator $D$ is trained over the real color image and the generated color image by $G$.
		The goal of $D$ is to distinguish real images from the fake ones.
		Both nets are trained adversarially.}
	\label{fig:gan_structure}
\end{figure}

GANs are generative models that learn a mapping from random noise vector $\bm{z}$ to an output color image $\bm{x}$:
$G : \bm{z} \rightarrow \bm{x}$.
Compared with GANs, conditional GANs learn a mapping from observed grayscale image $\bm{y}$ and random noise vector $\bm{z}$,
to $\bm{x}$: $G : \{\bm{y}, \bm{z}\} \rightarrow \bm{x}$.
The generator $G$ is trained to produce outputs that cannot be distinguished from ``real'' images by an
adversarially trained discriminator $D$, which is trained with the aim of detecting the ``fake'' images produced by the generator. This
training procedure is illustrated in Figure \ref{fig:gan_structure}.

The objective of a GAN can be expressed as
\begin{equation}
\begin{aligned}
\mathcal{L}_{\text{GAN}}(G,D) =&\mathbb{E}_{\bm{x}\sim P_{\text{data}}{(\bm{x})}}{[\log{D(\bm{x})}]}+\\
&\mathbb{E}_{\bm{z}\sim P_{\bm{z}}(\bm{z})}{[\log{(1-D(G(\bm{z})))}]},
\end{aligned}
\end{equation}
while the objective of a conditional GAN is
\begin{align}
\begin{aligned}
\mathcal{L}_{\text{cGAN}}(G,D) =& \mathbb{E}_{\bm{x}\sim P_{\text{data}}{(\bm{x})}}{[\log{D(\bm{x})}]}+ \\
&\mathbb{E}_{\bm{y}\sim P_{\text{gray}}(\bm{y}), \bm{z}\sim P_{\bm{z}}(\bm{z})}{[\log{(1-D(G(\bm{y},\bm{z})))}]},
\end{aligned}
\end{align}
where $G$ tries to minimize this objective against an adversarial $D$ that tries to maximize it,
i.e.
\begin{align}
G^* = \arg\min_{G}\max_{D}{\mathcal{L}_{\text{cGAN}}(G,D)}.
\end{align}

Without $\bm{z}$, the generator could still learn a mapping from $\bm{y}$ to
$\bm{x}$, but would produce deterministic outputs. That is why GAN is more suitable
for diverse colorization tasks than other deterministic neural networks.

\subsection{Architecture and implementation details}

The high-level structure of our conditional GAN is consistent with traditional ones \cite{Isola-et-al:CoRR-2016,Deshpande-et-al:CoRR-2016},
while the detailed architecture of our generator $G$ differs a lot.

\subsubsection{Convolution or deconvolution}
\begin{figure}[t]
	\centering
	\includegraphics[clip, trim={200pt, 0, 0, 220pt},
		width=0.85\textwidth]{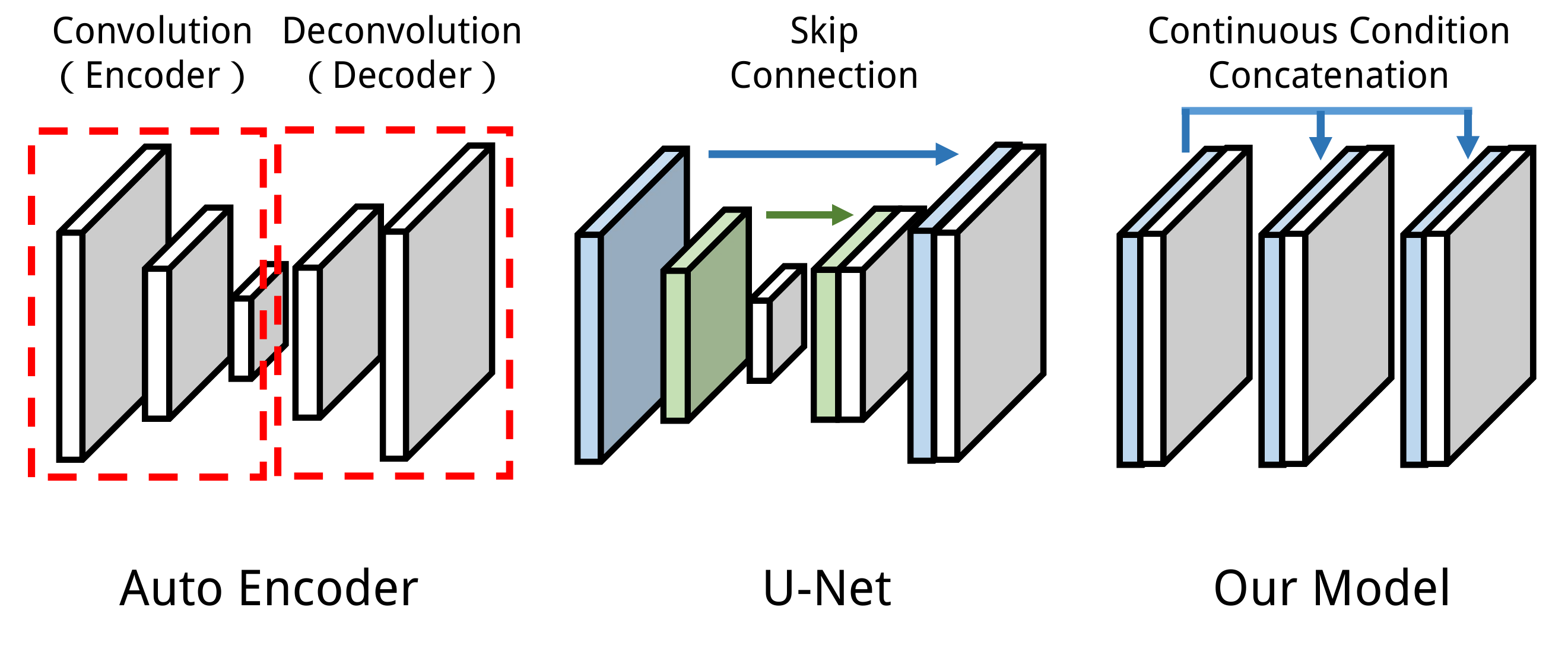}
	\vspace{-10pt}
	\caption{Structure comparison of auto encoder, U-Net and
		our generator. 
		Left: Auto encoder with convolutional encoder part and
		deconvolutional decoder part.
		Middle: U-Net structure \cite{Ronneberger-et-al:UNet-2015,Isola-et-al:CoRR-2016} with skip connections between
		layer $i$ and $n-i$.
		Right: Our fully convolutional generator with continuous
		condition concatenation}.
	\label{fig:UNet}
\end{figure}

Convolution and deconvolution layers are two basic components of image generators.
Convolution layers are mainly used  to exact conditional features.
And additionally, many researches \cite{Zhang-et-al:ECCV-2016,Isola-et-al:CoRR-2016,Deshpande-et-al:CoRR-2016}
use superposition of multiple convolution layers with stride more than 1 to downsize the data shape,
which works as a data encoder.
Deconvolution layers are then used to upsize the data shape as a decoder of the data representation \cite{Nguyen-et-al:DCNNCoRR-2016,Isola-et-al:CoRR-2016,Deshpande-et-al:CoRR-2016}.
While many other researches share this encoder-decoder structure, we choose to use only
convolution layers in our generator $G$. Firstly, convolution layers are well capable of feature extraction and transmission.
Meanwhile, all the convolution stride is set
to 1 to prevent data shape from downsizing, thus the important spatial information can be
kept along the data flow till the final generation layer.
Some other researches \cite{Ronneberger-et-al:UNet-2015,Isola-et-al:CoRR-2016} also takes this spatial information
into consideration. They add skip connections between each layer $i$
and layer $n - i$ to form a ``U-Net" structure, where $n$ is the total number of layers. Each
skip connection simply concatenates all channels at layer $i$ with those at layer $n - i$. 
Whether adding skip connections or not, the encoder-decoder structure more tends to extract global features and generate images by this overall
information which is more suitable for global shape transformation tasks. But in image colorization, we need a very detailed spatial local
guidance to make sure item boundaries will be accurately separated by different color parts in generated channels.
Let alone our modification is more straightforward and easy to implement.
See the structural difference between ``U-Net" and our convolution model in Figure~\ref{fig:UNet}.

\subsubsection{YUV or RGB}
A color image can be represented in different forms.
The most common representation is $RGB$ form which splits a color pixel into red, green, blue three channels.
Most computer vision tasks use $RGB$ representation \cite{Deshpande-et-al:ICCV-2015,Isola-et-al:CoRR-2016,Nguyen-et-al:VISIGRAPP-2016} due to its generality.
Other kinds of representations are also included like $YUV$ (or $YCrCb$) \cite{Cheng-et-al:ICCV-2015} and $Lab$ \cite{Charpiat-et-al:ECCV-2008,Zhang-et-al:ECCV-2016,Limmer-et-al:ICMLA-2016}.
In colorization tasks, we have grayscale image as conditional information, thus it is straightforward to use $YUV$ or $Lab$ representation
because the $Y$ and $L$ channel or so called Luminance channel represents exactly the grayscale information. So while using $YUV$
representation, we can just predict 2 channels and then concatenate with the grayscale channel to give a full color image.
Additionally, if you use $RGB$ as image representation, all result channels are predicted, thus to keep the grayscale of
generated color image consistent with the original grayscale image, we need to add an additional $L1$ loss
 as a controller to make sure $Gray(G(\bm{y},\bm{z})) \simeq \bm{y}$:
\begin{equation}
\label{equation:L1_loss}
\mathcal{L}_{L1}(G) = \mathbb{E}_{\bm{y}\sim P_{\text{gray}}(\bm{y}), \bm{z}\sim P_{\bm{z}}(\bm{z})}{[\| \bm{y}-Gray(G(\bm{y},\bm{z}))\|]},
\end{equation}
where for any color image $\bm{x} = (\bm{r},\bm{g},\bm{b})$, the corresponding grayscale image (or the Luminance channel $Y$) can be calculated by the well-known psychological formulation:
\begin{align}
Gray(\bm{x}) = 0.299\bm{r}+0.587\bm{g}+0.114\bm{b} .
\end{align}
Note that Eq.~(\ref{equation:L1_loss}) can still maintain good colorization diversity, because this $L1$ loss term only lays a constraint
on one dimension out of three channels.
Then the objective function will be modified to:
\begin{equation}
G^* = \arg\min_{G}\max_{D}{\mathcal{L}_{\text{cGAN}}(G,D)+\lambda\mathcal{L}_{L1}(G)}.
\end{equation}
Since there is no equality constraint between the recovered grayscale $Gray(G(\bm{y},\bm{z}))$ and the original one $\bm{y}$, the $\mathcal{L}_{L1}(G)$ factor will normally be non-zero, which
makes the training unstable due to this additional trade-off.
The results will be shown in Section \ref{section:Comparison} with experimental comparison on both $RGB$ and $YUV$ representations.


\subsubsection{Multi-layer noise}
The authors in \cite{Isola-et-al:CoRR-2016} mentioned noise ignorance while training the generator.
To handle this problem, they provide noise only in the form of dropout, applied on several layers of
the generator at both training and test time.
We also noticed this problem. Traditional GANs and conditional GANs receive noise information at the very
start layer, during the continuous data transformation through the network, the noise information is attenuated a lot.
To overcome this problem and make the colorization results more diversified, we concatenate the noise channel
onto the first half of the generator layers (the first three layers in our case).
We conduct experimental comparison on both one-layer noise and multi-layer noise representations, with
results shown in Section \ref{section:Comparison}.

\subsubsection{Multi-layer conditional information}
Other conditional GANs usually add conditional information only in the first layer,
because the layer shape of previous generators changes along their convolution and deconvolution layers.
But due to the consistent layer shape of our generator, we can apply concatenation of conditional
grayscale information throughout the whole generator layers which can provide sustained conditional supervision.
Though the ``U-Net" skip structure of \cite{Isola-et-al:CoRR-2016} can also help posterior layers receive conditional
information, our model modification is still more straightforward and convenient.

\subsubsection{Wasserstein GAN}
\begin{figure}[t]
	\centering
	\includegraphics[width=0.95\textwidth]{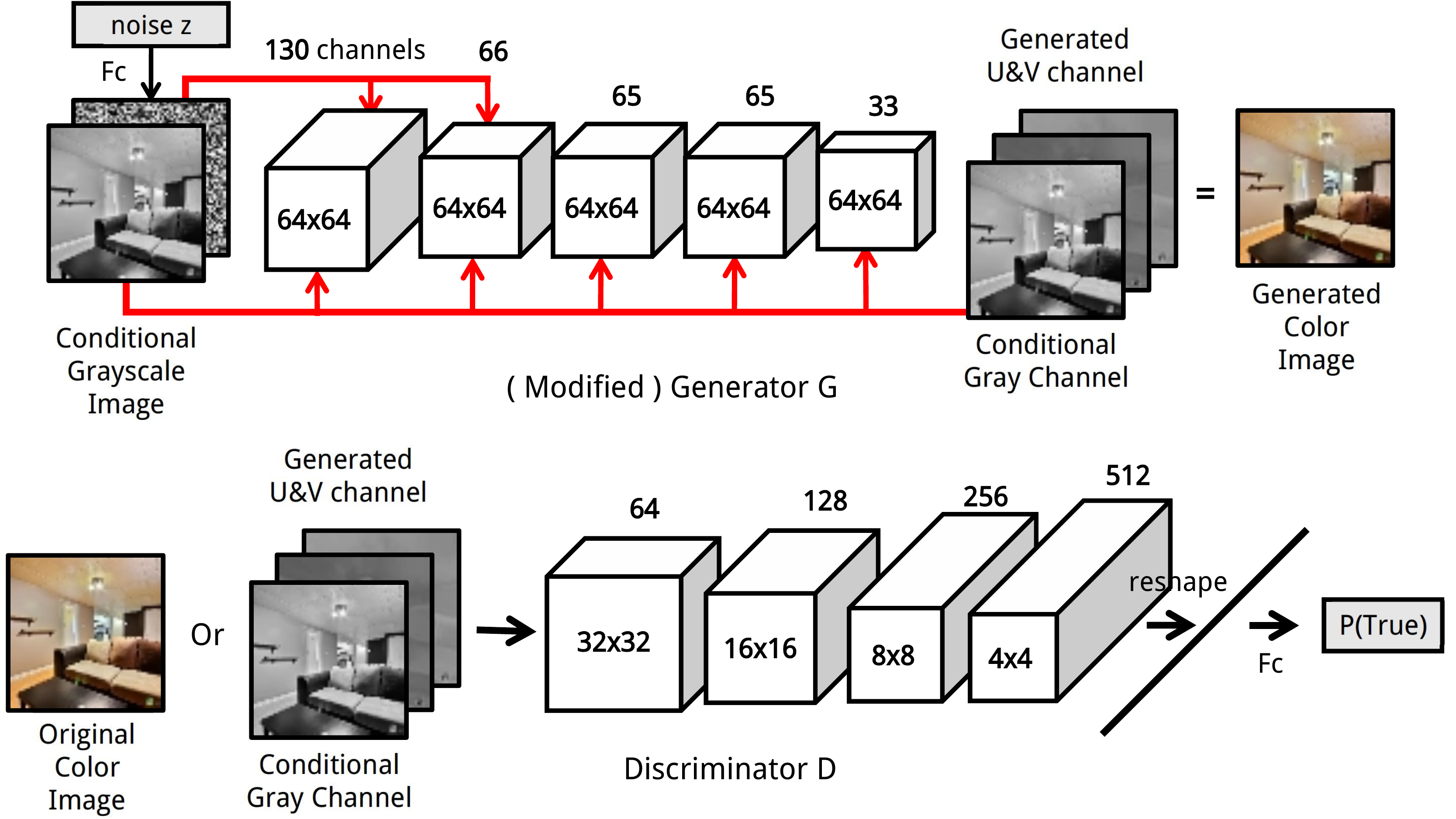}
	\caption{Detailed structure of our conditional GAN.
		Top: Generator $G$. Each cubic part represents a Convolution-BatchNorm-ReLU structure. Red connections
		represent our modifications of the traditional conditional GANs.
		Bottom: Discriminator $D$ }.
	\label{fig:detailed_structure}
\end{figure}

The recent work of Wasserstein GAN \cite{Arjovsky-et-al:WGAN2017} has acquired
much attention. The authors used Wasserstein distance to help getting rid of
problems in original GANs like mode collapse and gradient vanishing
and provide a measurable loss to indicate the progress of GAN training.
We also try implementing Wasserstein GAN modification into our model,
but the results are no better than our model. We make comparison between 
the results of Wasserstein GAN and our GAN in Section \ref{section:Comparison}.
\\
\\
The illustration of our model structure is shown in Figure~\ref{fig:detailed_structure}.

\subsection{Training and testing procedure}

\begin{algorithm}[t]
	\caption{Training phase of our conditional GANs, with the default parameters $k_D=1,k_G=1,m=64,s_z=100,s=64$.}
	\label{algorithm:training_phase}
	\begin{algorithmic}[1]
		\small
		\FOR{the number of training iterations}
		\FOR{$k_D$ steps}
		\STATE Generate minibatch of $m$ randomly sampled noise $\{\bm{z}^{(1)},\dots,\bm{z}^{(m)}\}$ each of size $[s_z]$.
		\STATE Sample minibatch of $m$ grayscale images $\{\bm{y}^{(1)},\dots,\bm{y}^{(m)}\}$
		each of shape $[s,s,1]$.
		\STATE Get the corresponding minibatch of $m$ color images $\{\bm{x}^{(1)},\dots,\bm{x}^{(m)}\}$
		from data distribution $p_{\text{data}}(\bm{x})$.
		\STATE Update the discriminator $D$ by:
		\[
		\nabla_{\theta_d}\frac{1}{m}\sum_{i=1}^{m}[\log{D(\bm{x}^{(i)})}+
		\log{(1-D(G(\bm{y}^{(i)},\bm{z}^{(i)})))}]
		\]
		\ENDFOR
		\FOR{$k_G$ steps}
		\STATE Generate minibatch of $m$ randomly sampled noise $\{\bm{z}^{(1)},\dots,\bm{z}^{(m)}\}$
		\STATE Sample minibatch of $m$ grayscale images $\{\bm{y}^{(1)},\dots,\bm{y}^{(m)}\}$
		\STATE Update the generator by:$$\nabla_{\theta_g}\frac{1}{m}\sum_{i=1}^{m}\log(1-D(G(\bm{y}^{(i)},\bm{z}^{(i)})))$$
		\ENDFOR
		\ENDFOR
	\end{algorithmic}
\end{algorithm}

The training phase of our conditional GANs is presented in Algorithm \ref{algorithm:training_phase}.
To assure the BatchNorm layers to work correctly, one cannot feed an image batch of the same images to test
various noise responses. Thus we use multi-round testing with same batch and rearrange them to test
different noise responses of each image, which is described in Algorithm \ref{algorithm:testing_phase}.

\begin{algorithm}[t]
	\caption{Testing phase of our conditional GANs with the default parameters $m=64,s_z=100$.}
	\label{algorithm:testing_phase}
	\begin{algorithmic}[1]
		\small
		\STATE Sample minibatch of $m$ grayscale images $\{\bm{y}^{(1)},\dots,\bm{y}^{(m)}\}$
		\FOR{round $i$ in $k_{\text{test}}$ rounds}
		\STATE Generate minibatch of $m$ randomly sampled noise $\{\bm{z}^{(1,i)},\dots,\bm{z}^{(m,i)}\}$ each of size $[s_z]$.
		\STATE Generate color image using the trained model $G$:
		\[
			\{\bm{x}^{(1,i)},\dots,\bm{x}^{(m,i)}\} \leftarrow G(\{\bm{y}^{(1)},\dots,\bm{y}^{(m)}\},\{\bm{z}^{(1,i)},\dots,\bm{z}^{(m,i)}\})
		\]
		\ENDFOR
		\STATE Rearrange generated results $\bm{x}$ into\\ $\{\bm{x}^{(1,1)},\dots,\bm{x}^{(1,k_{\text{test}})}\},\dots,\{\bm{x}^{(m,1)},\dots,\bm{x}^{(m,k_{\text{test}})}\}$
	\end{algorithmic}
\end{algorithm}

\section{Experiments}

\subsection{Dataset}

\begin{figure}[t]
	\centering
	\includegraphics[width=0.7\textwidth]{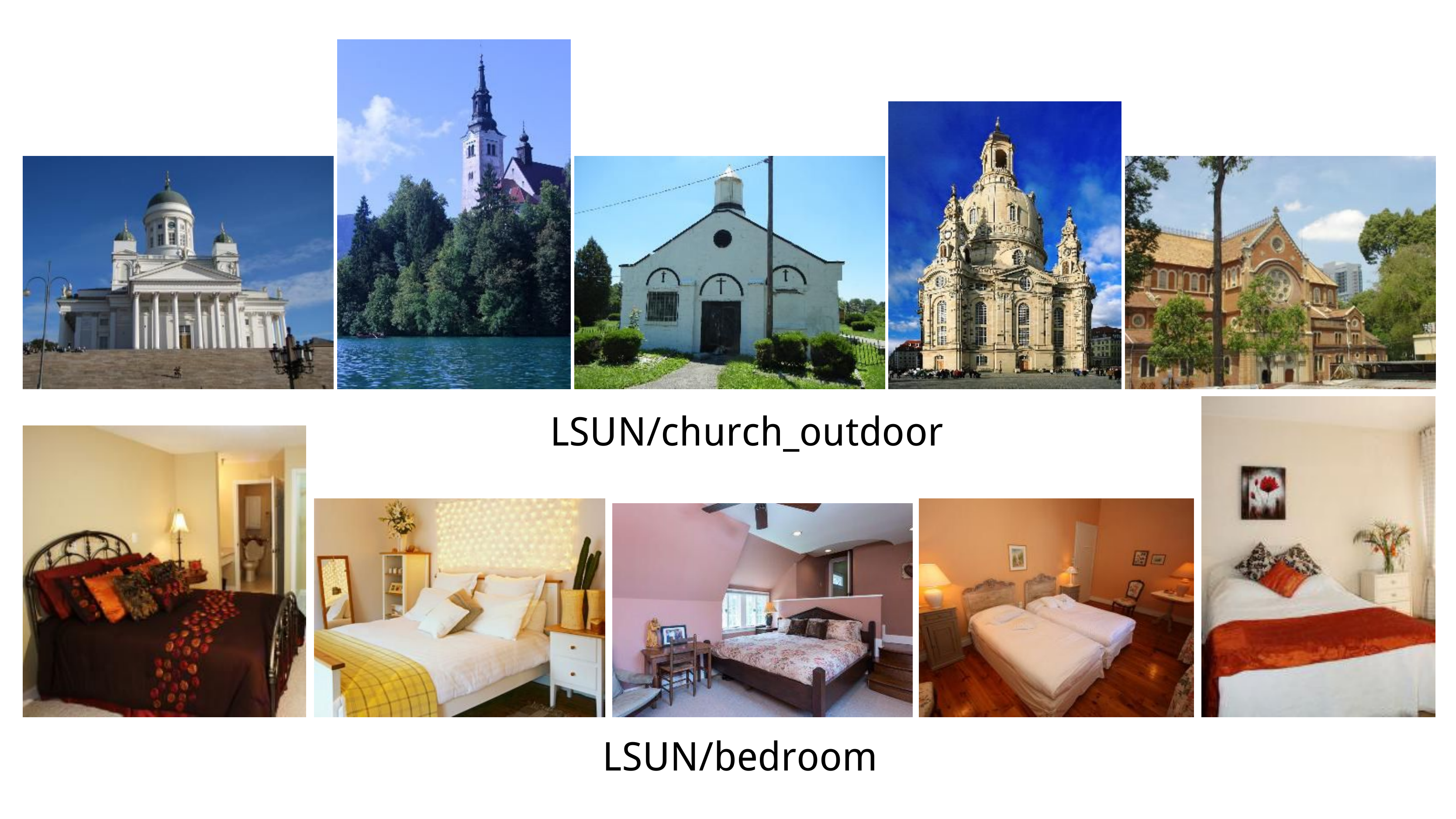}
	\vspace{-10pt}
	\caption{Demonstration of LSUN dataset.
		Top: Outdoor scene (church). Always blue sky and green trees.
		Bottom: Indoor scene (bedroom). Various item colors.}
	\label{fig:LSUN_demo}
\end{figure}

There are various kinds of color image datasets, and we choose the open LSUN bedroom dataset\footnote{LSUN dataset is available at \url{http://lsun.cs.princeton.edu}.} \cite{Yu-et-al:LSUN-2015} to conduct our experiment.
LSUN is a large color image dataset generated iteratively by human labeling with automatic deep neural classification.
It contains around one million labeled images for each of 10 scene categories and 20 object categories.
Among them we choose indoor scene bedroom because it has enough samples (more than 3 million) and unlike outdoor scenes
in which trees are almost always green and sky is always blue, items in indoor scenes like bedroom have various colors, as shown in Figure \ref{fig:LSUN_demo}.
This is exactly what we need to fully explore the capability of our conditional GAN model.
In experiments, we use $503,900$ bedroom images randomly picked from the LSUN dataset.
We crop a maximum center square out of each image and reshape them into $64\times64$ pixel as
preprocessing.

\subsection{Comparison Experiments}
\label{section:Comparison}

\subsubsection{YUV and RGB}
\begin{figure}[t]
	\centering
	\includegraphics[clip, trim={440pt, 0, 0, 0},
		width=0.65\textwidth]{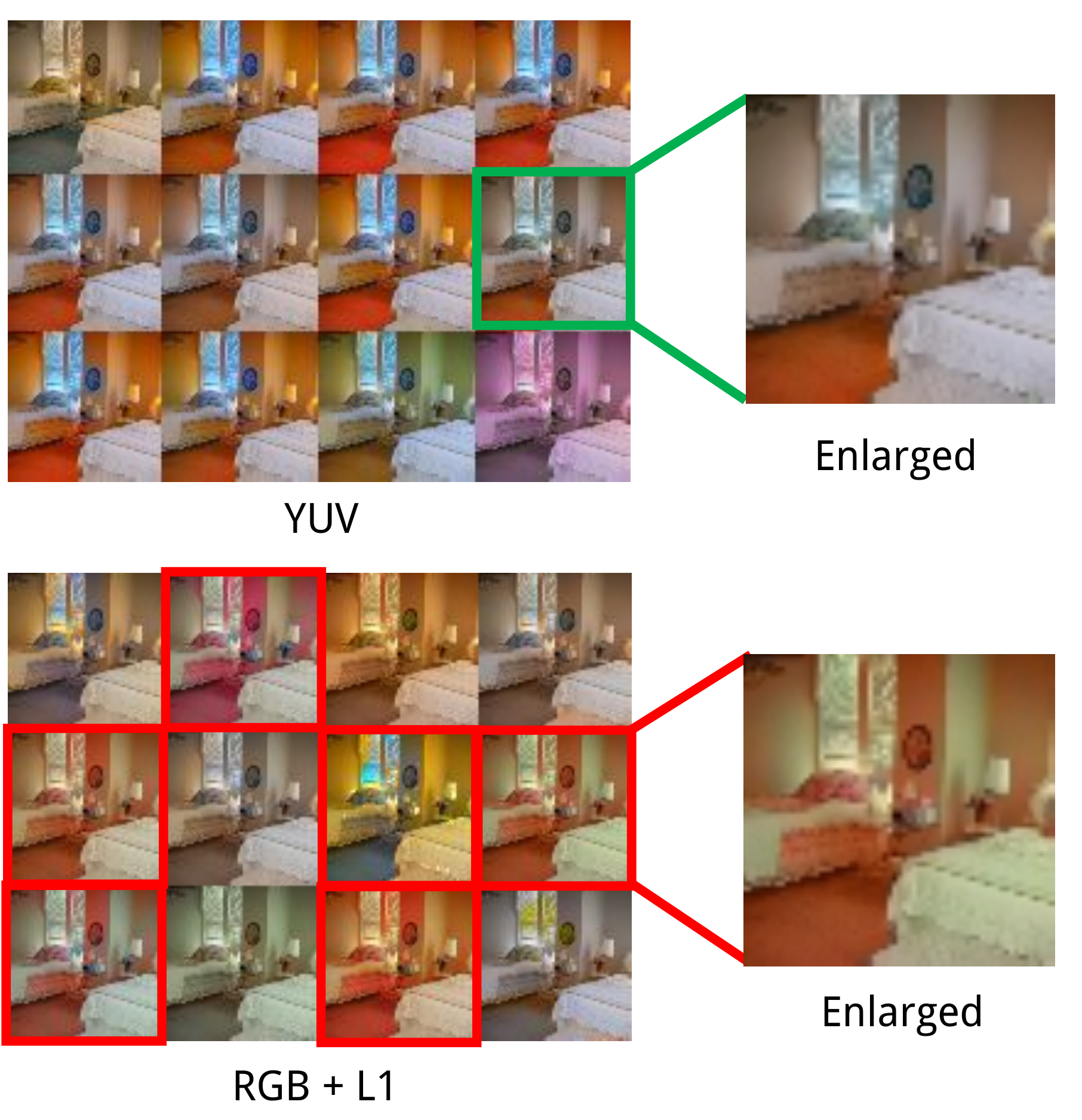}
	\vspace{-10pt}
	\caption{Comparison of different color space representation.
		Top: training and testing with $YUV$ representation.
		Bottom: training with $RGB$ representation and $L1$ loss.
		Focus on the results in red boxes, $RGB$ representation results
		lack of item continuity due to the additional trade-off between $L1$ loss and the GAN loss.}
	\label{fig:YUV_RGB_result}
\end{figure}

The generated colorization results of a same grayscale image using $YUV$ representation and $RGB$ representation
with additional $L1$ loss are shown in Figure \ref{fig:YUV_RGB_result}.
Each group of images consists of $3\times4$ images generated from a same
grayscale image by each model at a same epoch.
Focus on the results in red boxes, we can see $RGB$ representation suffers from structural miss due to the additional trade-off between $L1$ loss and the GAN loss. Take the enlarged image on the top right in Figure \ref{fig:YUV_RGB_result} as an example, both the wall and the bed on 
the left are split by unnaturally white and orange colors, 
while the results of $YUV$ setting acquire more smooth transitions. Moreover, the model using $RGB$ representation shall predict 3 color channels while $YUV$ representation only predicts 2 channels given the
grayscale $Y$ channel as fixed conditional information,
which makes the $YUV$ model training much more stable.

\subsubsection{Single-layer and multi-layer noise}
\begin{figure}[t]
	\centering
	\includegraphics[width=0.65\textwidth]{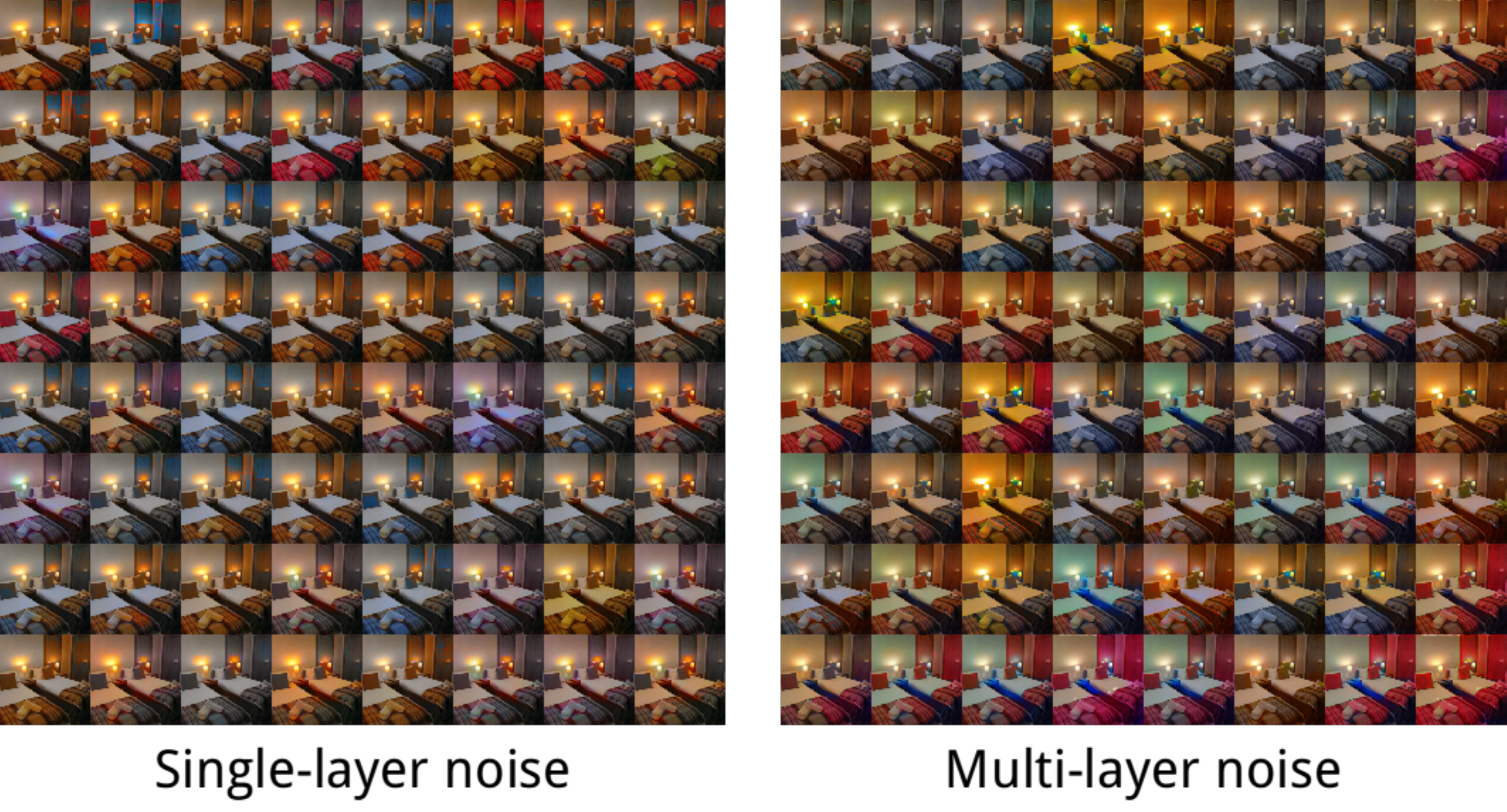}
	\vspace{-10pt}
	\caption{Comparison of single-layer and multi-layer noise model results.
		Left: results of single-layer noise model.
		Right: results of multi-layer noise model.
		Apparently multi-layer noise possesses our generator $G$ of higher diversity.}
	\label{fig:multi_noise_result}
\end{figure}

The generated colorization results of the same grayscale images using single-layer noise model and
multi-layer noise model are shown in Figure \ref{fig:multi_noise_result}.
Each group consists of $8\times8$ images generated from a grayscale image by each model at the same epoch.
We can see from the results that multi-layer noise possesses our generator $G$ of 
higher diversity as those results on the right in Figure \ref{fig:multi_noise_result} are apparently
more colorful.
\begin{figure}[h]
	\centering
	\includegraphics[clip, trim={540pt, 0, 0, 135pt}, 
		width=0.65\textwidth]{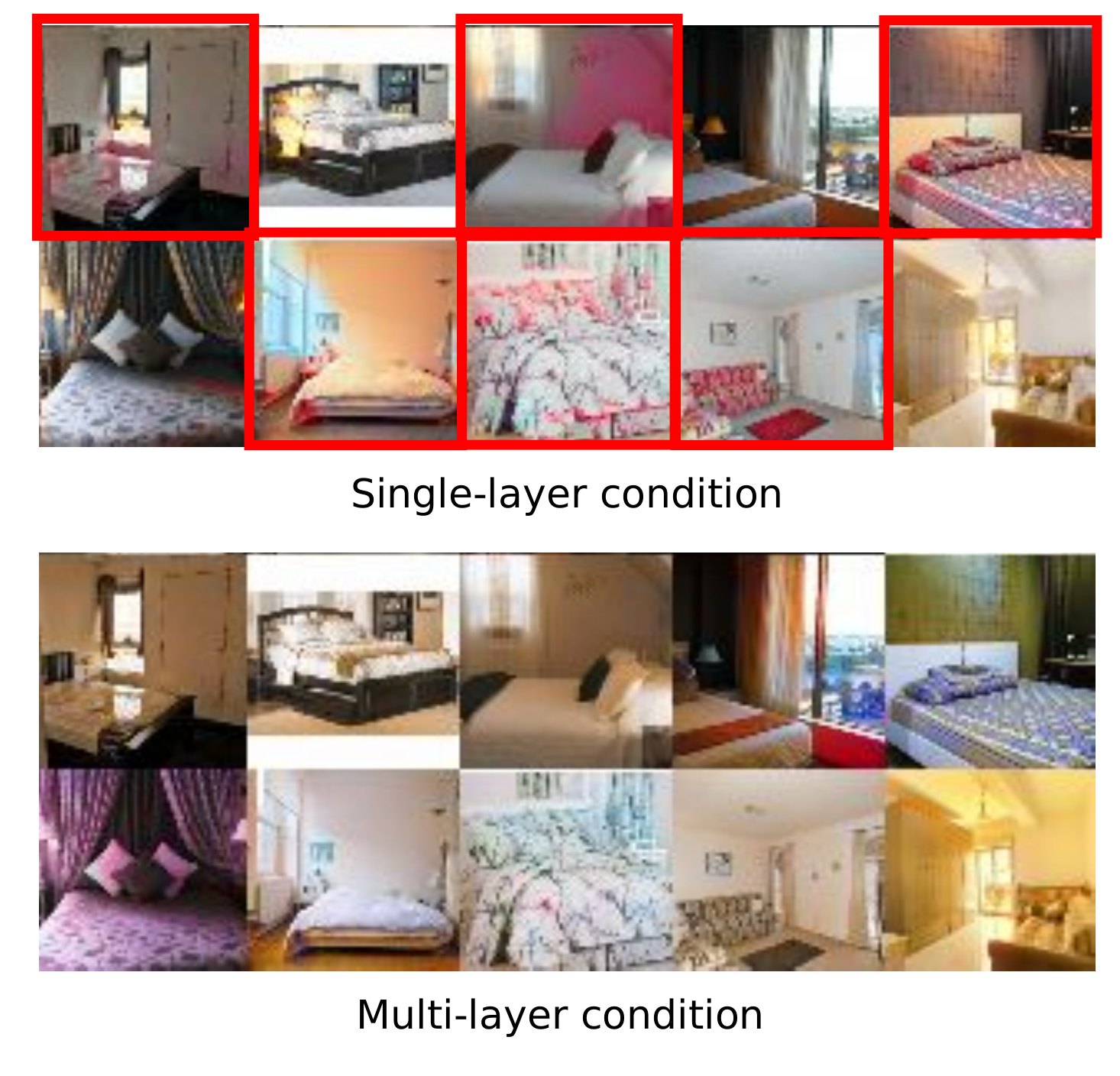}
	\vspace{-15pt}
	\caption{Comparison of single-layer and multi-layer condition model result.
		Top: results of single-layer condition model, suffer from colorization derivation (red box).
		Bottom: results of multi-layer condition model, smooth transition.}
	\label{fig:continuous_condition_result}
\end{figure}

\subsubsection{Single-layer and multi-layer condition}
The generated colorization results of the same grayscale images using single-layer conditional model and
multi-layer conditional model are shown in Figure \ref{fig:continuous_condition_result}.
We show $2\times5$ images generated by single-layer condition setting
and multi-layer condition setting at the same epoch.
We can see from the results that the multi-layer condition model makes the generator more structural information
and thus the results of multi-layer condition model are more stable while the single-layer conditional model
suffers from colorization derivation like those images in red boxes
in Figure \ref{fig:continuous_condition_result}.

\subsubsection{Wasserstein GAN}
\begin{figure}[t]
	\centering
	\includegraphics[clip, trim={185pt, 0, 0, 150pt}, 
		width=0.95\textwidth]{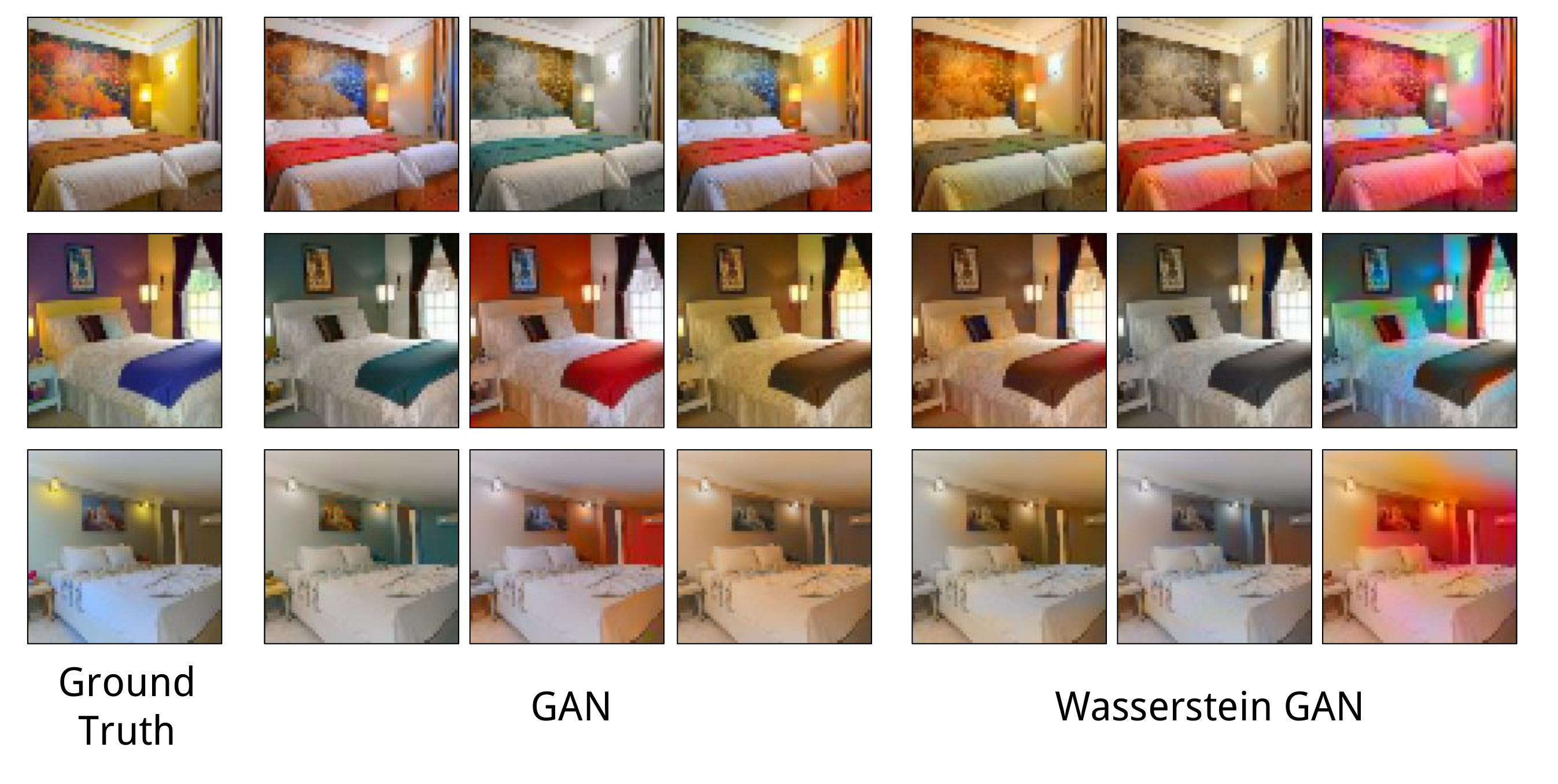}
	\vspace{-12pt}
	\caption{Comparison between the results of GAN and Wasserstein GAN.
		Each line consists of the leftmost ground truth color image
		and three results by GAN, then three results by Wasserstein GAN.
		The rightmost images are failed results by Wasserstein GAN.}
	\label{fig:wgan_result}
\end{figure}

Three groups of colorization results of the same grayscale images using GAN
and Wasserstein GAN are shown in Figure \ref{fig:wgan_result}.
We can see from the result that Wasserstein GAN can produce comparable
results as the first two column of Wasserstein GAN shows, but there are still failed results by Wasserstein GAN like the last column, even to
note that the Wasserstein GAN results(40 epoch) come after training twice the number of epochs of the GAN results(20 epoch).
This is mainly due to that our training LSUN bedroom dataset is quite large (503,900 image),
the discriminator will not overfit easily, which prevents the gradient
vanishing problem. And also because of the large dataset, 
the discriminator needs quite a lot times of optimization to
convergence, not to mention Wasserstein GAN shall not use momentum 
based optimization strategies like Adam due to the non-linear
parameter value clipping, Wasserstein GAN needs much longer training
to produce comparable results as our model. Since Wasserstain GAN
only helps to improve the stability of GAN training at a price of
much longer training time and we have
achieved results of good reality through our GAN, we did not use
Wassserstein GAN structure.

More results and discussion of our final model will be shown in the next section.

\section{Results and Evaluation}

\subsection{Colorization Results}
\label{section:result}

\begin{figure}
	\centering
	\includegraphics[width=0.98\textwidth]{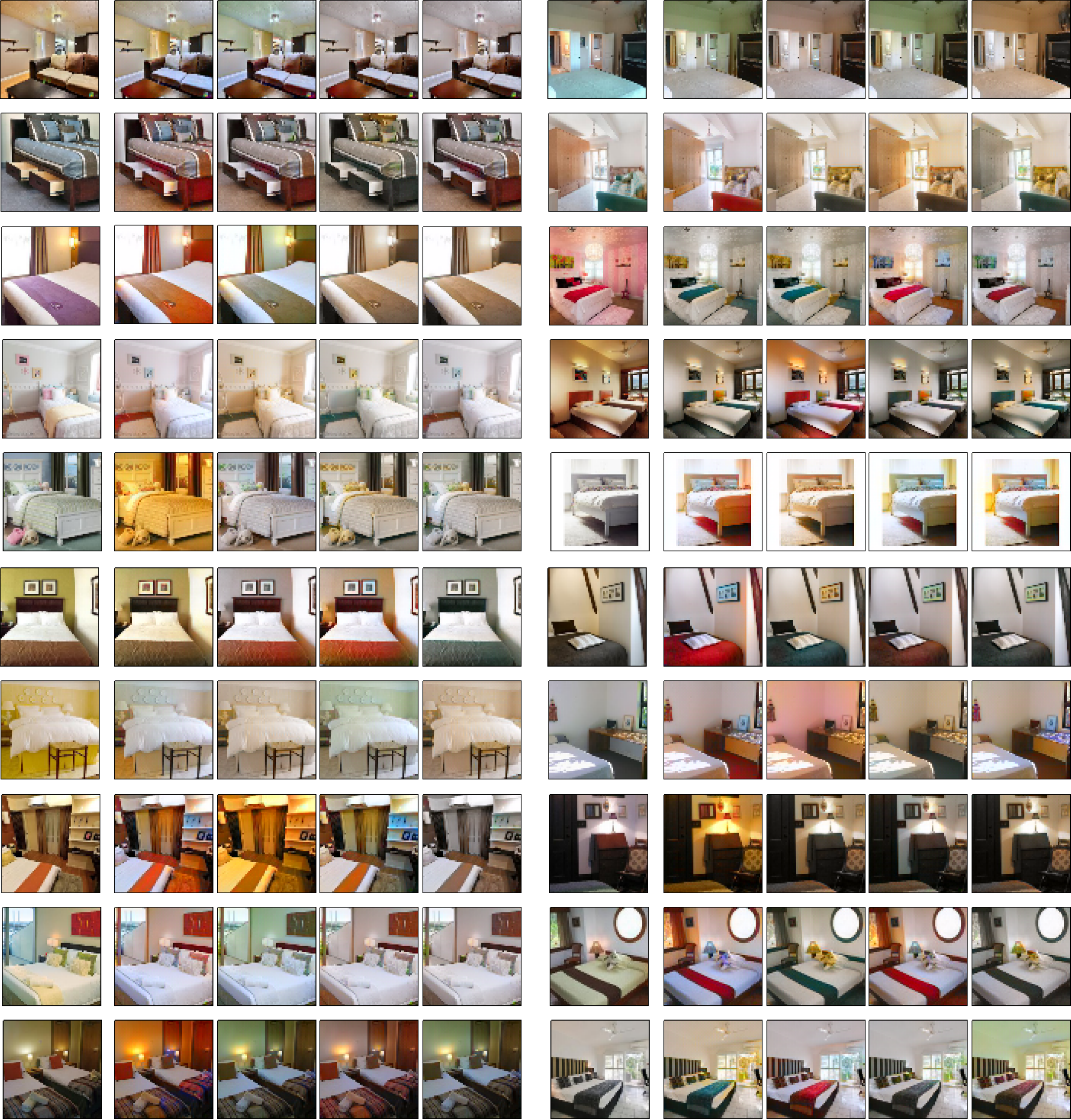}
	\vspace{8pt}
	\caption{Example results of our conditional GAN on LSUN bedroom data.
		20 groups of images, each consists of the \textbf{leftmost} ground truth color image and 4 different
		colorizations generated by our conditional GANs given the grayscale version of the ground truth image.
		One can clearly see that our novel structure
		generator produces various colorization schemes while maintaining
		good reality.}
	\label{fig:final_result}
\end{figure}

A variety of image colorization results by our conditional GANs are provided in Figure \ref{fig:final_result}.
Apparently, our fully convolutional (without stride) generator with multi-layer noise and multi-layer condition concatenation
produces various kinds of colorization schemes while maintaining
good reality. Almost all color parts are kept within correct components without deviation.

\subsection{Evaluation via Human Study}

Previous methods share a common goal to provide a color image closer to the original one.
That is why many of their models \cite{Jung-et-al:JSC-2016,Perarnau-et-al:CoRR-2016,Deshpande-et-al:CoRR-2016}
take image distance like RMSE (Root Mean Square Error) and PSNR (Peak Signal-to-Noise Ratio)
as their measurements.
And others \cite{Iizuka-et-al:ACM-2016,Isola-et-al:CoRR-2016} use additional classifiers to predict if colorized image
can be detected or still correctly classified.
But our goal is to generate diverse colorization schemes, so we cannot take those distance as our measurements
as there exist reasonable colorizations that diverge a lot from the original color image.
Note that some previous work on image colorization \cite{Cheng-et-al:ICCV-2015,Dong-et-al:CoRR-2017,Nguyen-et-al:VISIGRAPP-2016,Nguyen-et-al:DCNNCoRR-2016}
does not provide quantified measurements.

Therefore, just like some previous researches \cite{Isola-et-al:CoRR-2016,Zhang-et-al:ECCV-2016},
we provide questionnaire surveys as a Turing test to measure our colorization results.
We ask each of the total 80 participants 20 questions. In each question, we display 5 color images, one of which is the ground truth image, the others are our
generated colorizations of the grayscale image of the ground truth, and ask them if any one of them is of poor reality.
We add ground truth image among them as a reference in case of participants do not think any of them is real.
And we arrange all images randomly to avoid any position bias for
participants.
The feedback from 80 participants indicates more than 62.6\% of our generated color images are convincible while the rate of ground truth images is 70.0\%.
Furthermore, we run significance $t$-test between the ground truth and the generated images on the percentages of humans rating as real image for each question. The $p$-value is $0.1359>0.05$, indicating our generated results have no significant difference with the ground truth images.
Also we calculate the credibility rank within each group of the ground truth image and the four corresponding generated images. An image gets
higher rank if higher percentage of participants mark it real.
And the average credibility rank of the ground truth images is only 2.5
out of 5, which means at least $(2.5-1)/(5-1)=37.5\%$ of our generated results are even more convincible than true images.

\section{Conclusion}
In this paper, we proposed a novel solution to automatically
generate diverse colorization schemes for a grayscale image while maintaining their reality by exploiting conditional generative
adversarial networks which not only solved the sepia-toned problem of other models but also enhanced the colorization diversity. 
We introduced a novel generator architecture which consists of fully convolutional non-stride structure with multi-layer noise to enhance
diversity and multi-layer condition concatenation to maintain reality.
With this structure, our model successfully generated diversified high-quality color images for each input grayscale image.
We performed a questionnaire survey as a Turing test to evaluate our colorization result.
The feedback from 80 participants indicates our generated colorization results are highly convincible.

For future work, as so far we have investigated methods to generate color images by conditional GAN given only corresponding grayscale images, which provides the model maximum freedom
to generate all kinds of colors, we can also lay additional
constraints on the generator to guide the colorization procedure.
Those conditions include but not limited to 
(i) specified item color, such as blue bed and white wall etc.; and 
(ii) global color scheme, such as warm tone or cool tone etc.
And note that given those constraints, generative adversarial networks shall still produce various vivid colorizations.

\clearpage

\clearpage

\appendix

\bibliographystyle{splncs03}
\bibliography{colorGAN}

\end{document}